# Using Multi-modal Data for Improving Generalizability and Explainability of Disease Classification in Radiology

Pranav Agnihotri[1], Sara Ketabi[2,3], Khashayar (Ernest) Namdar[2,5,7], and Farzad Khalvati[2,3,4,5,6,7]

[1]Engineering Science, University of Toronto, Toronto, ON, Canada

[2]The Hospital for Sick Children (SickKids), Toronto, ON, Canada

[3]Department of Mechanical and Industrial Engineering, University of Toronto, Toronto, ON, Canada

[4]Department of Medical Imaging, University of Toronto, Toronto, ON, Canada

[5]Institute of Medical Science, University of Toronto, Toronto, ON, Canada

[6]Department of Computer Science, University of Toronto, Toronto, ON, Canada

[7]Vector Institute, Toronto, ON, Canada

Abstract

Traditional datasets for the radiological diagnosis tend to only provide the radiology image alongside the radiology report. However, radiology reading as performed by radiologists is a complex process, and information such as the radiologist's eye-fixations over the course of the reading has the potential to be an invaluable data source to learn from. Nonetheless, the collection of such data is expensive and time-consuming. This leads to the question of whether such data is worth the investment to collect. This paper utilizes the recently published Eye-Gaze dataset to perform an exhaustive study on the impact on performance and explainability of deep learning (DL) classification in the face of varying levels of input features, namely: radiology images, radiology report text, and radiologist eye-gaze data. We find that the best classification performance of X-ray images is achieved with a combination of radiology report free-text and radiology image, with the eye-gaze data providing no performance boost. Nonetheless, eye-gaze data serving as secondary ground truth alongside the class label results in highly explainable models that generate better attention maps compared to models trained to do classification and attention map generation without eye-gaze data.

## 1. Introduction

The rapid success of deep learning (DL) has resulted in its applications in various domains including medical imaging and radiology. The consequences of poor performance on medical tasks are vast, and thus it is no wonder that better model performance is desirable. As with most other DL tasks, the success of applying neural network (NN) approaches to classification in radiology depends highly on the data available. However, the field of medical imaging suffers greatly from a data shortage, with many research groups only having a few hundred or thousand data points to work with from local hospitals [1][2].

There exist some publicly available chest X-ray datasets in the order of hundreds of thousands of data points that can be used for training and testing DL models. Examples of such datasets are CheXpert, and MIMIC-CXR [3][4]. However, these datasets are not very feature-rich compared to the sources of information present during a live radiology reading. These datasets, at best, tend to only provide the radiology images, patient demographic information, and their associated radiology reports as data points. From a reductionist perspective, these are indeed the key elements that comprise the inputs and outputs of a radiology reading. However, a crucial source of information ignored is the actual process of the reading that the radiologist does. This involves information about their eye movements and what points of the image they fixate on.

Such sources of information become even more important to consider in the context of explainability. There have been multiple examples of DL methods showing great promise in medical imaging, even matching the performance of human practitioners [5][6][7]. However, the adoption of DL methods has been slow because of the inability of healthcare professionals to understand the decisions made by these "black-box" models [8]. A common method of providing explainable models is attention map generation. Attention maps show which parts of an image a NN has focused on in order to make its classification. Traditionally, models generate such maps using the weights of their network being trained for a task such as pathology classification [9]. The goal of the model is not in-and-of-itself to generate accurate maps, but the hope is that if it learns the proper underlying patterns for the classification task, the attention maps will automatically be more precise and useful. In this context, having the eye-gaze information of a radiologist is invaluable as a data source to learn from.

Yet, aggregating eye movement data is challenging. Collecting eye-tracking information requires new equipment, and the cost of implementing new infrastructure in hospitals and clinics to facilitate digitization of medical information such that they can be accessed by researchers can be quite significant. Moreover, the construction of datasets for a task with the help of a radiologist suffers from issues with scaling. Apart from the time/cost of performing readings, the process of generating the radiology report involves recording the audio of a radiologist dictating their findings, and that audio being sent to a transcriber. The radiologist then reviews the transcription and makes edits if needed [10]. Thus, building an extensive dataset from scratch is difficult and costly.

In order to justify such costs, there must be evidence that this new data source improves the performance of DL approaches. There exists a gap in the literature in the quantification of performance gains that come by adding different levels of context in radiology classification. The recently created Eye Gaze Dataset by A. Karargyris et al. [11] provides information about a radiologist's eye gaze as well as the radiology image and report. The eye-gaze data appears in two forms: temporal and static. The temporal data is a sequence of heatmaps showing where a radiologist fixated upon an image over the course of the reading, and the static data is an amalgamation of the temporal data. The pixel intensity in the heatmaps corresponds to the time of the radiologist's focus on the pixel.

Apart from the added sources of information, this dataset contrasts existing ones in that it provides ground truth for three classification labels, namely Normal, Congestive Heart Failure (CHF), and Pneumonia, which comes from a group of interdisciplinary clinicians [11]. Pre-existing datasets rarely do this, and researchers have to mine the reports for key words that can indicate the label associated with the radiology image [12]. This severely constricts the ability to use the full radiology report as an input to the model, as it could lead to circular reasoning given that the label too comes from the report. The lack of this problem in this Eye Gaze dataset opens the possibility of fully exploring the potential information packed in radiology reports.

Thus, it is now possible - using the Eye Gaze dataset - to thoroughly address the aforementioned gap in the literature, and evaluate performance and explainability differences that come from using simply the radiology image as input to a model, versus appending other data modalities such as radiology report free text, and the radiologist's eye-gaze information.

## 2. Background Work

A. Karargyris et al. - the creators of the Eye Gaze dataset - provide a baseline framework for multi-class classification on 1083 chest X-ray images using only the radiology image as input during training [11]. The radiology image is resized to a dimension of 224×224 pixels and then passed through a Unet [13] encoder that funnels into a linear layer to classify the three aforementioned disease labels, i.e., Normal, CHF, and Pneumonia.

In order to introduce the radiologist's eye-gaze information to the model, the authors propose two methods. The first method makes use of the temporal eye-gaze data in the dataset. The temporal heatmaps are converted into a sequence of vectors via a Convolutional Neural Network (CNN) encoder, and those sequence vectors are passed through a bi-directional Long Short-Term Memory (LSTM) [14]. Separately, the X-ray image is encoded into a vector by means of a CNN encoder. The image vector is concatenated with the output of the LSTM, and together pass through a linear layer for the final classification.

The second method of introducing eye-gaze data to the model is to treat it as secondary ground truth alongside the class label, rather than use it as an input feature. The architecture proposed is that of a Unet, similar to the baseline model. At the Unet bottleneck layer, one branch of the model continues onwards to pass through linear layers to do the prediction. Another branch of the model is upsampled to output an attention map. It is with this attention map that the secondary loss is calculated by means of the static heatmap ground truth. Thus, this is a multi-loss architecture where the eye gaze information supplements the training of the model by acting as secondary ground truth to the class label.

Across the baseline model, temporal model, and static heatmap models discussed above, the authors reported an area under the receiver operating characteristic curve (AUC) of 0.87, 0.81, and 0.87, respectively. However, as these results were not obtained through cross-validation, more rigorous results should be confirmed by training models on different folds of the dataset. Moreover, while the paper does provide a framework for how to utilize the eye-gaze information, it does not experiment with the radiology report text and establish its usefulness as an input feature.

The above method of using a heatmap as ground truth is called heatmap regression, and is widely used for the problem of pose estimation in computer vision [15]. The authors of [16] show how CNN architecture can use deconvolution layers to generate heatmaps that can be compared against target heatmaps using Mean Squared Error (MSE) loss. Within the medical field, heatmap regression has been used to detect breast mass detection in mammography by Zheng et al. [17]. They used a fully convolutional network (FCN) that took in 439 mammography images as input and outputted a heatmap. For labels, they converted the weak annotations they had into heatmaps and used an F-score loss function to compare the predicted heatmap versus ground truth heatmap.

This method of heatmap regression seems naturally suited as a way to incorporate radiologist eye-gaze information into a model.

In a similar method, Pesce et al. utilize heatmaps as ground truth to localize areas in more than 430,000 chest radiographs [18]. Their ground truth is a highly localized heatmap such that when overlapped with the radiology image, it only focuses on the area where the lesion indicative of the pathogen is, and not the rest of the image. They pass the input image through convolutional layers and extract saliency maps from high levels of those layers. The saliency maps serve to calculate the model's inferred position in the image where the lesions indicative of the pathogen is. This predicted location is compared to the ground truth location, and a localization error is backpropagated. This approach is adaptable to situations where it is clear the radiologist is highly focused on specific points, and rather than calculating loss via MSE or F-score across the entire generated heatmap, a localization loss between ground truth points and predicted points is utilized.

So far, the focus has been on the radiology image and heatmaps stemming from a radiologist's eye gaze information. Another significant source of information is the radiology reports themselves. To that end, there are several notable methods of performing classification involving radiology reports.

Shin et al. propose a method of performing classification tasks using solely the radiology report as an input to the model [19]. The classification task tackled in their paper was not that of pathology classification, but rather severity classification and presence of certain conditions such as intracranial bleeding or acute stroke. They first compiled a corpus consisting of all the radiology head computed tomography reports they had access to outside of the dataset they would train their model on. The corpus is tokenized, and word embeddings are trained by the original implementation of word2vec [20] using Continuous Bag of Words (CBOW) and SKIP models and negative sampling. They found a word embedding size between 200 to 400 to work the best. They report that creating their own word embeddings on documents from their task domain showed improvement in performance rather than relying on word embedding methods such as Term Frequency (TF) or Term Frequency-Inverse Document Frequency (TF-IDF) [21]. After creating the word embedding model, they use it to convert each radiology report in their dataset into a document matrix by generating a vector for all the non stop words. The document matrix is fed into a simple CNN architecture that ends with a dense vector. Before this dense vector can be passed through a final linear layer that performs the classification, it is appended with a global embedding attention vector. This global embedding attention vector is formed by taking the document matrix and applying a 1×1 convolution to it. The result of the 1×1 convolution is called the attention matrix, and by performing a max-pool on each row of the attention matrix, an attention vector is formed. At this point, the original document matrix is transposed and multiplied by the attention vector to finally get the global embedding attention vector. This is the vector that is appended to the output of the CNN. Finally, the penultimate vector is passed into a classification layer. For the classification tasks, they report accuracy values averaging 89%. While the task itself is not the exact same as pathology classification, the method of word-embedding creation and attention generation is transferable.

The idea of creating global attention (or saliency) values is a popular technique, and was utilized by Wang et al. who proposed a method to integrate radiology report free text and radiology images for classifying 3,643 chest X-ray images into three classes, namely Pneumothorax, Pleural Effusion, and Mass. They proposed a CNN-Recurrent Neural Network (RNN) framework called TieNet that takes as input a chest X-ray image, and generates a disease classification, a small radiology report, as well as a heatmap [9]. To generate a heatmap over the input image, a spatial attention map is created at each timestep of the RNN. In order to generate a final heatmap, the authors use saliency weighted global average pooling. They reuse the RNN's attention mechanism, picking out the maximum attention head vector value at each RNN timestep, to generate a sequence of saliency values. These saliency values are used to weight each timestep's spatial attention map, and the weighted sum of all spatial attention maps is convoluted with the CNN's final transition layer feature map to generate the final heatmap. The authors propose that this heatmap is an improved visualization because it uses information from both the CNN and the RNN, and thus represents learnings from both data sources. The authors also showed how TieNet could discard report-generation and simply map reports and images to a classification label by only backpropagating the classification loss. They commented on how radiology reports had rich, easy-to-learn features and so helped improve the classification accuracy. Thus, attempts to improve radiology classification scores could benefit from leveraging both report free-texts and images. However, the primary focus of this paper was to generate a report after training. Although they saw some success, aiming to reproduce such highly variable and comparatively lengthy radiology reports (on the order of several sentences spanning several radiology-report sections) resulted in inconsistent outputs that were of poor quality [22].

In response to the poor-quality generated reports, Gale et al. proposed generating shorter text sequences using a CNN-LSTM model on 50,363 frontal pelvic X-rays, which consists of 4,010 hip fractures [22]. Their goal was to output a short sentence for an input X-ray image, describing the location and character of a fracture. First, they finetuned a pre-trained DenseNet model for performing hip fracture detection (classification into fracture or no fracture). They then took the final activation map from the CNN and fed it into an LSTM with attention. The LSTM was trained to generate short sentences using a set of radiologist's hand-labeled descriptive terms for the corresponding image. Although the authors had access to free-text radiology reports just as the authors of [9] did, they showed that training text generation on long, unstructured text was quite difficult, and thus had a radiologist condense the reports into the short descriptive sentences that resulted in a much higher BiLingual Evaluation Understudy (BLEU) [23] score during evaluation. However, although the authors of [22] saw better text-generation performance than the TieNet model of [9], their scope was limited by the fact that they were not performing a multi-disease classification task (a fracture was the only abnormality that could be present in an image). As well, [9] suffered from the issue of data-modality: only one image was used for prediction, and no other metadata was leveraged.

Rodin et al of [12] attempt to solve the data-modality issue above by using a similar architecture to TieNet, except instead of only using radiology images as input variables, they used the radiology report as well. Their data includes 8,530 frontal antero-posterior chest X-ray images corresponding to a random subset of the MIMIC-CXR database. It is worth noting though that the MIMIC-CXR dataset used in their study does not come with ground truth classification labels. It consists of radiology images and the corresponding free-text reports. Labels were mined using natural language processing (NLP) techniques from the report. Radiology reports are structured to have several sections, each providing different information. Of relevance to this study are the History section, Findings section, and Impressions section. The History section contains information about a patient's age, sex, and anamnesis (account of medical history). The age and sex information were extracted with regular expressions and turned into categorical variables to feed into the model. Details about patient medical history were extracted with text-processing techniques, abbreviations were unabbreviated using a dictionary, and then converted into word embeddings using word2vec pretrained on PubMed to serve as the second data input stream. The Findings and Impressions sections of a report hold information about the radiologist's diagnosis and any abnormalities or pathogens detected. This information was mined using NLP techniques and squeezed into short sentences similar to [22] that had the structure "[pathology],[present/absent], [(optional) location], [(optional) severity]". This served as the label against which the RNN was trained to output sequences. Thus, with image and report, the authors were able to produce short diagnostic sentences. The authors also propose an extension to their model borrowed from TieNet [9] in order to solve the localization and classification problem. For each RNN timestep, they extracted the most salient attention values which are used to do a weighted sum of the spatial attention map also generated at each timestep. The saliency values are also used to weight the text embeddings at each time step. At the end, the final weighted heatmap and text-embeddings are fed into a linear layer to produce a classification, similar to [9]. However, in order to do this extension, one needs a ground-truth label for the classification, which their data did not include. Thus, they had to test it against a separate dataset that had images and labels, but no radiology reports. The complete effectiveness of leveraging report and image to solve classification, localization, and image captioning were thus not fully explored. Nonetheless, their overall contribution is an architecture that can perform classification, localization, as well as producing short text describing the image, using radiology reports and images as the input.

## 3. Methods

### 3.1. Data

The MIMIC-CXR database provides a vast repository of radiology images and reports [4]. The Eye Gaze dataset [11] as a subset of this database, provides the following features: 1,083 chest X-ray images, the accompanying radiology reports, and the recorded eye-gaze information of a radiologist performing a reading on the X-ray images. The eye-gaze information appears in temporal form and static form. The temporal form is a sequence of heatmaps, each showing a snapshot of the intensity with which the radiologist was focusing on different parts of the image, where intensity is proportional to the time spent examining the position. The static form is an amalgamation of the temporal heatmaps. The dataset also provides ground truth classification labels spanning three classes: Normal, CHF, and Pneumonia. The ground truth labels come from external medical practitioners, not the radiologists who wrote the reports or performed the eye-gaze readings. During the initial data processing phase, it was found that 66 of the chest X-rays were missing completed eye-gaze information. Thus, this study makes use of the 1,017 data points which are not missing any information.

Furthermore, a radiology report consists of several sections, three notable ones being the Exam Indication section, the Findings section, and the Impressions section. The Findings and Impressions sections of a radiology report entail the diagnosis of a radiologist after they have done their reading, whereas the Exam Indication section is a short medical history of patient symptoms and why they appear for a radiology exam to begin with. Word-embedding vectors were determined using 30,000 reports mined from the MIMIC-CXR database. There is no overlap between these reports and the reports available in the Eye-Gaze dataset. It ensures that the models trained on the related word-embedding features will be evaluated based on a completely separate dataset from what it has been trained on, preventing potential biases.

We aim to examine the effect of radiology reports on our models with and without the involvement of radiologists. Thus, we first train our models on the images and the Indication section of the reports. Then, to evaluate whether the radiologist's findings can help the framework and also whether the image features can improve what radiologists find, we use the entire reports along with the images in a separate experiment.

An example of a report from the dataset illustrates the difference between the two:
- Exam Indication
    - *Evaluation of the patient with fever first day after repair of umbilical hernia*
- Full Report
    - *Evaluation of the patient with fever first day after repair of umbilical hernia. PA and lateral upright chest radiographs were reviewed in comparison to ___. Right pleural effusion is partially loculated, moderate associated with bibasal atelectasis. No appreciable pneumothorax is seen. No new consolidations demonstrated, but the consolidations in the lung bases might potentially represent at least in part infectious process*

Thus, the key sources of information from the Eye Gaze dataset to train a model to perform radiology classification are:
1) Chest X-ray
2) Radiology report text
    a) Exam Indication
    b) Full Report
3) Eye-gaze information
    a) Temporal Heatmaps
    b) Static Heatmap

As the goal is to measure the impact of different features on classification performance and explainability, two sets of experiments are carried out; one for performance, and the other for explainability.

## 3.2. Classification Performance Experiments

Given the multitude of data sources to incorporate into the model, the first step is to quantify how much information is encoded in each feature independently. Then, the combinations of input features are tested. The main evaluation metric for this study is the mean AUC and AUC for each class, which has been calculated based on the One-vs-Rest approach. In other words, for calculating the per-class AUC, we considered three binary problems for each label, and calculated the AUC based on the presence or absence of that label corresponding to a datapoint. The results for each experiment will be discussed in Section 4.

### 3.2.1. Chest X-ray

The architecture used for performing classification with just the chest X-ray as input is in line with the baseline model proposed by the authors of the Eye Gaze dataset [11]. It is visualized in Figure 1. The chest X-rays were resized to 224×224 before being passed into the model.

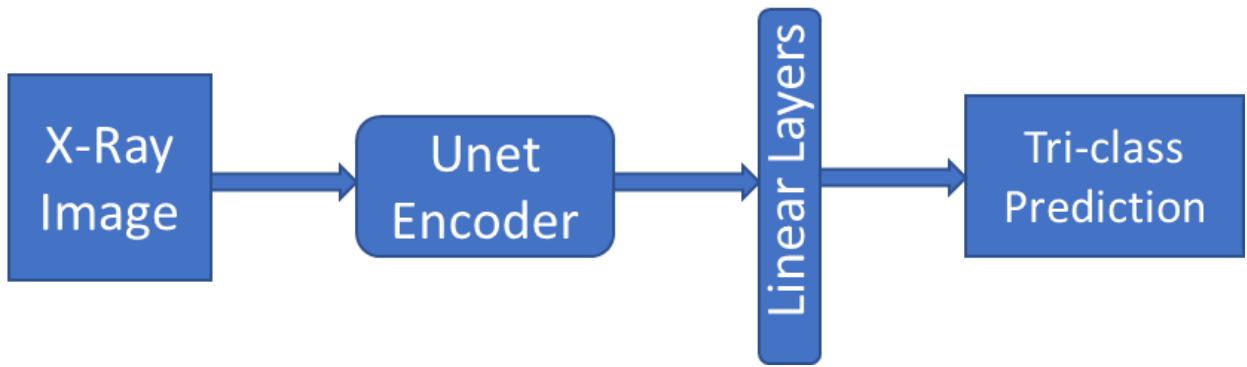

*Figure 1: Unet Encoder architecture for X-ray input classification*

### 3.2.2 Eye-gaze information

We incorporated the eye-gaze information through static and temporal heatmaps in two separate experiments. The static heatmap was passed into the same architecture as the chest X-ray, while the temporal heatmaps were fed into a CNN-RNN architecture. The architectures used for the two experiments are visualized in Figures 2 and 3, respectively.

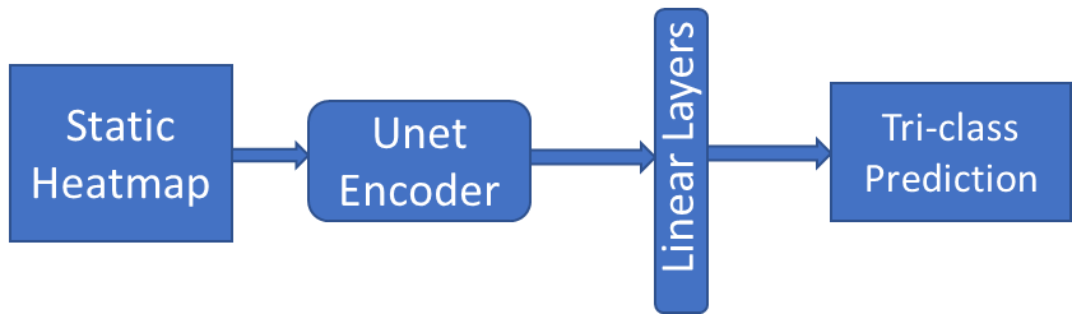

*Figure 2: Architecture for static heatmap input classification*

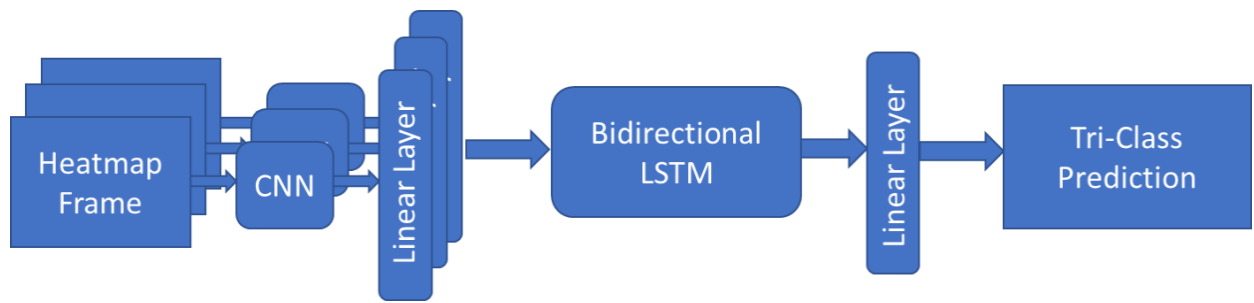

*Figure 3: CNN-RNN architecture for temporal heatmap input classification*

### 3.2.3 Radiology Report

In order to work with natural language, a word embedding model pretrained on radiology reports was found to work the best as compared to off-the-shelf biological embedding models such as BioBert [24]. Word-embeddings of size 150 were trained using a Word2Vec Skip-gram scheme on the extracted MIMIC CXR reports. Each word in the input text was cast to its corresponding word vector. As well, an *average embedding* was calculated simply by averaging the word vectors for every vocabulary word in the model. Words in the input text that were out of vocabulary were given the value of this average embedding. In testing, it was discovered that good classification performance could be achieved by simply creating a sentence embedding for the entire input text by summing up the individual word vectors.

The same model architecture was used for both the exam indication and full report experiments, which can be found in Figure 4. A sentence embedding was created for the whole Exam Indication input. For the full report input, however, every word in the report was considered as the input to the trained word embedding model.

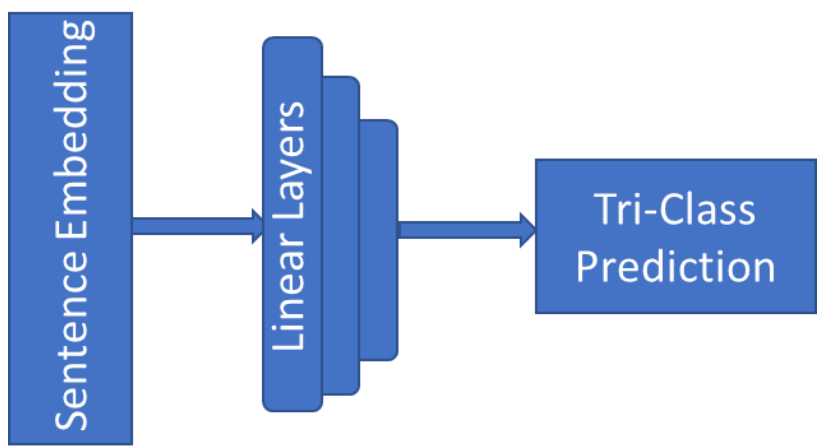

*Figure 4: Architecture for classification with text embedding input*

In order to understand why full report text is so highly predictive as an input feature, the sentence embeddings were visualized for both Exam Indication text, as well as Full Report text. Principal Component Analysis (PCA) was used to project the 150-dimensional sentence embedding vectors down to their first 2 principal components. The two-dimensional points were then plotted. Exam Indication sentence embeddings are plotted in Figure 5-a, and Full Report sentence embeddings are plotted in Figure 5-b. The plot legend of 1,2,3 corresponds to pathogen classes Normal, CHF, and Pneumonia, respectively.

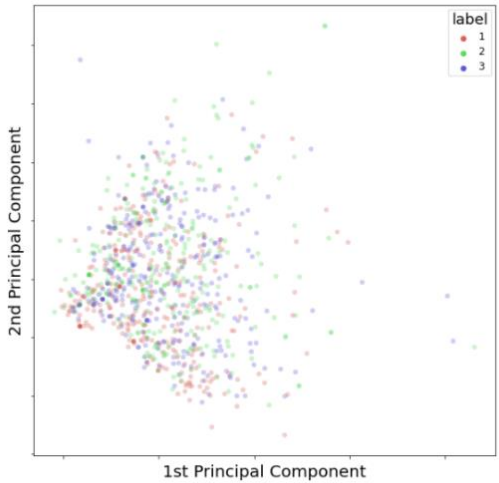 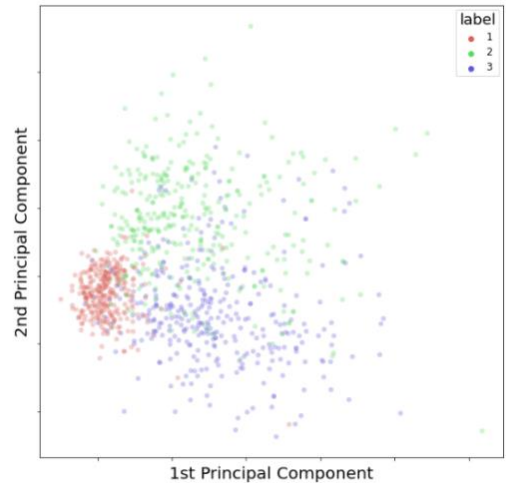

*Figure 5-a*                          *Figure 5-b*

*Figure 5: a) Exam Indication sentence embeddings visualized in 2 dimensions. b) Full Report sentence embeddings visualized in 2 dimensions*

Although there is no discernible pattern for the Exam Indication sentence embeddings, there is a clear class-based clustering for the Full Report embeddings. This implies that it is the textual content in the Findings and Impressions section of the report (the content not available in Exam Indication text) that is defining the class boundaries. It also means that for the task of Normal vs CHF vs Pneumonia classification, the radiologists are quite accurate with their diagnosis. However, this result may not be extensible to other tasks such as cancer detection, which has a lower radiologist accuracy rate. In such situations, the reliance on features influenced by radiologist biases and mistakes may not be desirable.

3.2.4 Radiology Report Text and Chest X-Rays

The architecture used to combine either the Full Report sentence embedding and the chest X-ray image or the Exam Indication sentence embedding, and the image is visualized in Figure 6. The sentence embeddings for Exam Indication and full report inputs were obtained as discussed in the previous experiments. The outputs of the image and sentence embeddings features were combined before the final classification layer, and this layer was applied to the fused features.

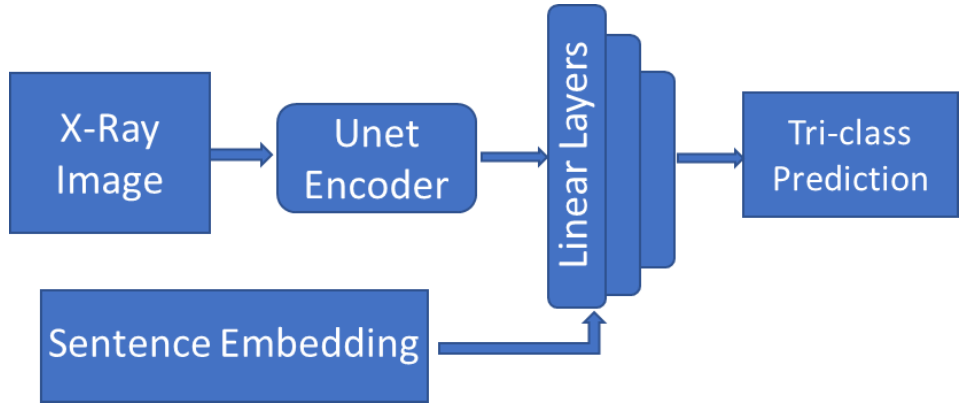

*Figure 6: Architecture for the classification model taking in chest X-ray and sentence embedding input*

3.2.5 Eye-Gaze Information and Chest X-Rays

Eye-gaze information as a source of learning for the classification model can be included in two ways, as the authors of the Eye-Gaze dataset showed [11]: as ground truth or as an input. Through experiments, it was found that the most effective use of the static eye-gaze heatmap was as ground truth against which a secondary mask loss could be computed. Therefore, the incorporation of this data source into the framework can make the model learn how to generate it. The architecture is shown in Figure 7. The total loss was formed by combining the two losses with the weightage given by the Eye-Gaze dataset paper authors, as per their hyperparameter tuning report [11].

$$total\ loss = (0.5827 \times \alpha) + (0.4173 \times \beta)$$

Where $\alpha$ is the heatmap loss and $\beta$ is the classification loss.

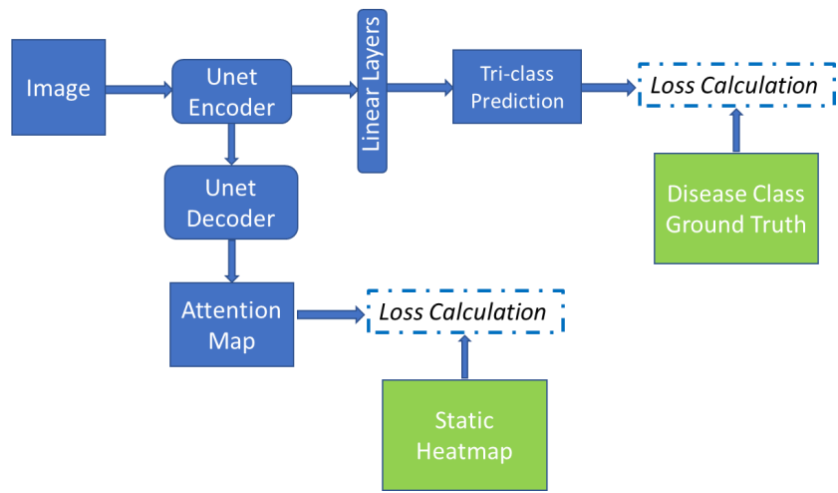

*Figure 7: Architecture for chest X-ray input with static heatmap ground truth*

The temporal eye-gaze heatmaps can be combined with chest X-ray images as a fused input. Therefore, they were first converted into a sequence of vectors via a CNN encoder, and those sequence vectors were then passed through a bi-directional LSTM. Separately, the X-ray image was encoded into a vector by means of a Unet encoder. The image vector was concatenated with the output of the LSTM, and together passed through a linear layer for the final classification. This is shown in Figure 8.

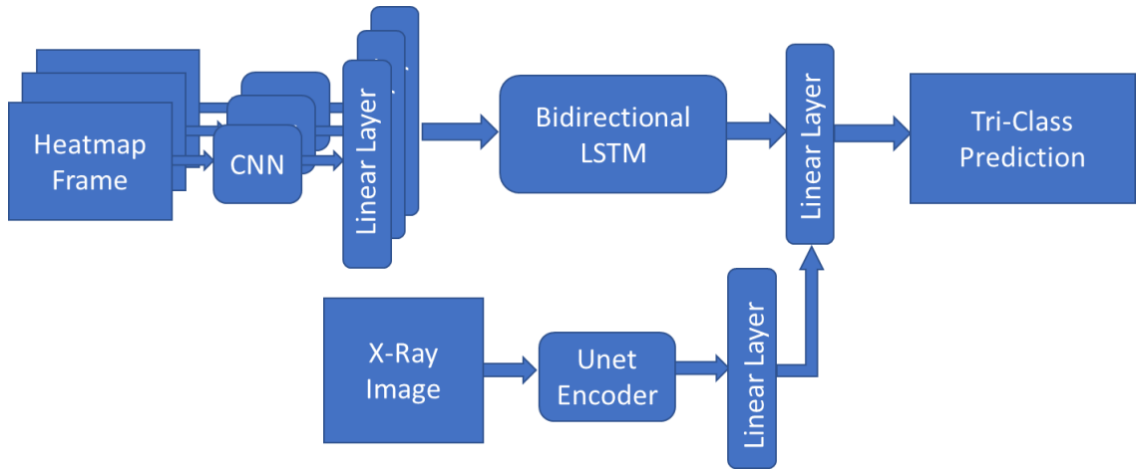

*Figure 8: Architecture for temporal heatmaps and chest X-ray input to classifier*

3.3 Explainability Experiments

The mode of explainability chosen for this study is the generation of attention maps by the classification model to highlight areas in the input chest X-ray image relevant to the predictions. In order to judge the quality of the generated attention maps, the MIMIC-CXR Annotations dataset is used [25]. It consists of 350 chest X-rays diagnosed with Pneumonia, along with the radiologist generated bounding boxes highlighting areas of interest. Each image is also accompanied by a radiology report.

The quality of the generated attention map is calculated by measuring the intensity of the attention map within the bounding box, normalized by the intensity of attention over the rest of the image. The pixel values of the attention map lie in the range of 0 to 255. We defined the metric as Eq 1.

$$Attention\ overlap = \frac{\sum_{x \in B} \emptyset(x)}{\sum_{x \in I} \emptyset(x)} \quad (1)$$

where,

$$\emptyset(x) = \begin{cases} x & x > 100 \\ 0 & x \leq 100 \end{cases}$$

and $B$ is the set of pixels within the bounding boxes and $I$ is the set of all pixels in the image.

The cut-off of pixel intensities at a value of 100 serves to create a stricter metric that requires the pixels of the attention map that lie inside the bounding boxes to have a bare-minimum level of intensity in order to contribute to the metric. By use of this metric, relative comparisons were made between models that learn to generate attention maps from different features.

The tested feature combinations are:
- Chest X-ray (baseline)
- Chest X-ray and Exam Indication text
- Chest X-ray and Full Report text

Since the models are outputting an attention map in addition to a classification, eye-gaze data can be introduced to the models by using the static heatmaps as secondary ground truth to train the attention map generation. In addition to the classification loss, a second loss is computed between the generated map and the static heatmap. Thus, for each of the above three feature combination models, attention map generation will be learned with and without the static heatmap ground truth. This allows evaluation on the effect of eye-gaze data on explainability.

Figure 9 depicts an example of an X-ray image containing bounding box annotations along with the corresponding generated eye-gaze static heatmaps. In this example, the mean attention overlap is 0.241.

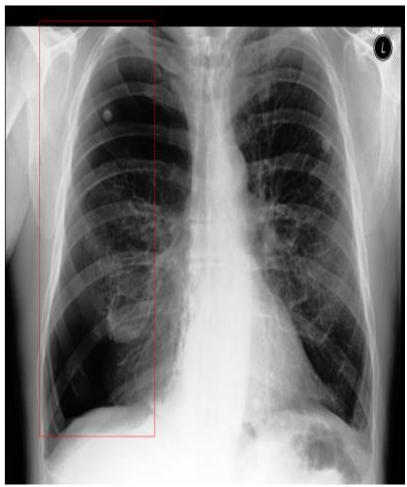 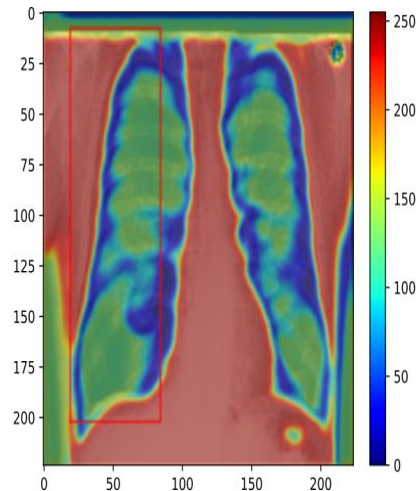

*Figure 9-a*          *Figure 9-b*

*Figure 9: a) Instance of bounding box annotations from the MIMIC-CXR Annotations dataset and b) The corresponding generated static heatmap*

The used architectures for the models without and with the static heatmap as the second loss are visualized in Figures 10 and 7, respectively. The attention map outputted from the Unet decoder is used to compute the attention overlap metric.

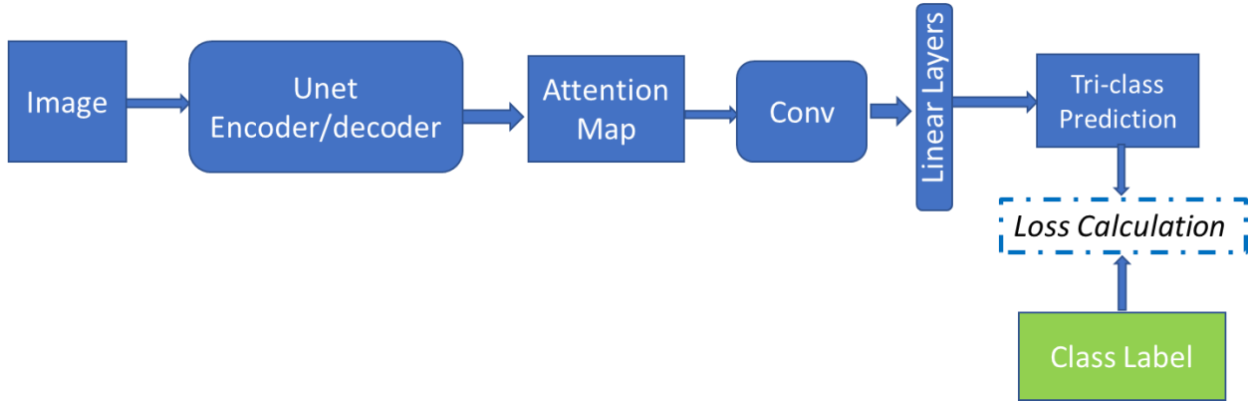

*Figure 10: Architecture to generate attention map with chest X-ray input*

We investigated the explainability of the model with the chest X-ray input combined with either the indication part of the report or the full one. The same architecture was used for both experiments, which is depicted in Figure 11. For the situations where the static heatmap is not being used as the secondary loss, the heatmap loss branch will not be active. In these models, the attention map is only learned through the classification loss being backpropagated.

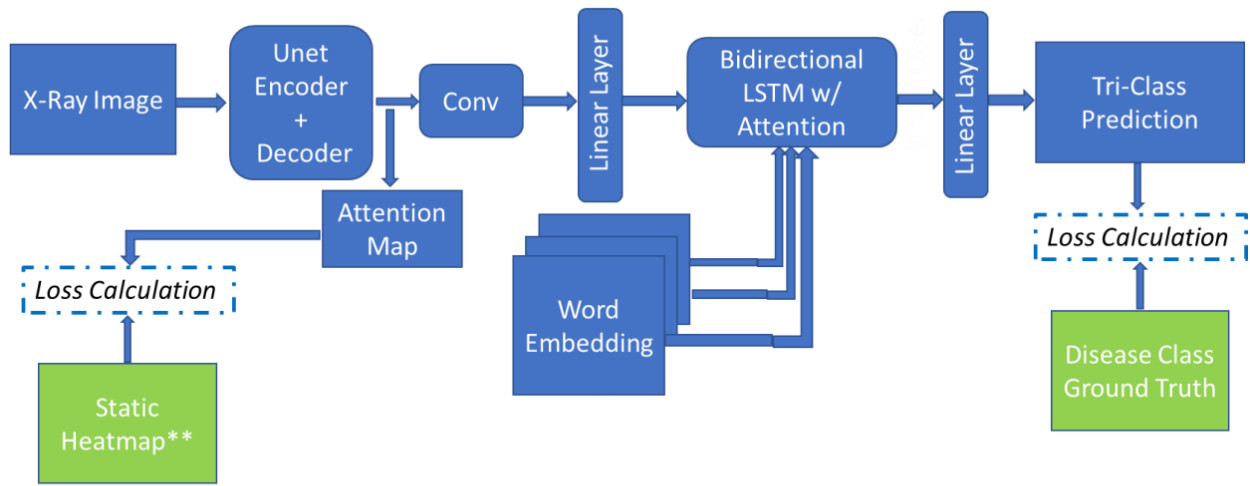

*Figure 11: The CNN-RNN model architecture used for all image and word embedding experiments*

For each of the aforementioned models, the mean attention overlap will be used as a measure of explainability. In other words, the higher overlap between the bounding boxes and the important areas of the generated attention maps indicates that the model's attention has been on relevant areas of the image, meaning that it will be more explainable to radiologists. The obtained mean attention overlap values will be represented in the Results section.

# 4. Results

In this section, we will explain the results of the experiments discussed in Section 3. First, the obtained classification performance based on the average overall AUC and per-class AUC will be represented. The results are stated based on 5-fold cross-validation. To prevent the potential bias of spreading a patient ID into different folds, we held the same patient IDs across the folds of the experiments. Then, we will examine the explainability of the trained models by showing the mean and median values for the attention overlap between the generated attention maps and ground-truth bounding boxes.

4.1. Classification Performance Results

4.1.1 Chest X-ray

The results of 5-fold cross-validation of the model with only X-ray input using the architecture shown in Figure 1 can be found in Table 1.

*Table 1: Classification AUC with chest X-ray input*

|  | **Fold 1** | **Fold 2** | **Fold 3** | **Fold 4** | **Fold 5** | **Average** |
|---|---|---|---|---|---|---|
| **Normal** | 0.835 | 0.868 | 0.911 | 0.921 | 0.842 | **0.875** |
| **CHF** | 0.808 | 0.906 | 0.931 | 0.968 | 0.907 | **0.904** |
| **Pneumonia** | 0.723 | 0.807 | 0.757 | 0.84 | 0.821 | **0.789** |
| **Average AUC** | **0.788** | **0.86** | **0.866** | **0.911** | **0.857** | **0.856** |

Note that the AUC for the Pneumonia class lags behind the other two classes. It seems to be a class that the model using visual inputs struggles to classify.

4.1.2 Static Heatmap

The results of 5-fold cross-validation of the model with only static heatmap input using the architecture shown in Figure 2 can be found in Table 2.

*Table 2: Classification AUC with static heatmap input*

|  | **Fold 1** | **Fold 2** | **Fold 3** | **Fold 4** | **Fold 5** | **Average** |
|---|---|---|---|---|---|---|
| **Normal** | 0.816 | 0.824 | 0.871 | 0.804 | 0.841 | **0.832** |
| **CHF** | 0.721 | 0.734 | 0.643 | 0.694 | 0.812 | **0.721** |
| **Pneumonia** | 0.684 | 0.686 | 0.814 | 0.661 | 0.695 | **0.707** |
| **Average AUC** | **0.741** | **0.748** | **0.776** | **0.719** | **0.783** | **0.753** |

From Table 2, it is evident that the classification performance drops significantly with only the eye-gaze static heatmap as input, at least compared to the model that took in the chest X-ray images as input. However, it is still outperforming random classifiers, thus indicating that there is indeed some discriminative power in the eye-gaze data such that the three classes can be differentiated to an extent.

4.1.3 Temporal Heatmap

The results of 5-fold cross-validation of the model with only temporal heatmap input using the architecture shown in Figure 3 can be found in Table 3.

*Table 3: Classification AUC with temporal heatmap input*

|  | **Fold 1** | **Fold 2** | **Fold 3** | **Fold 4** | **Fold 5** | **Average** |
|---|---|---|---|---|---|---|
| **Normal** | 0.722 | 0.881 | 0.889 | 0.765 | 0.762 | **0.803** |
| **CHF** | 0.751 | 0.691 | 0.739 | 0.746 | 0.704 | **0.726** |
| **Pneumonia** | 0.551 | 0.699 | 0.665 | 0.539 | 0.631 | **0.616** |
| **Average AUC** | **0.674** | **0.757** | **0.764** | **0.683** | **0.699** | **0.715** |

From Table 3, it is seen that although the classification performance drops compared to the static heatmap input, the signs of some discriminative power are still evident in that the model performs better than a random classifier.

4.1.4 Radiology Report: Exam Indication

The results of 5-fold cross-validation of the model with only the exam indication part of the reports as input using the architecture shown in Figure 4 can be found in Table 4.

*Table 4: Classification AUC with exam indication text input*

|  | *Fold 1* | *Fold 2* | *Fold 3* | *Fold 4* | *Fold 5* | *Average* |
|---|---|---|---|---|---|---|
| *Normal* | 0.832 | 0.811 | 0.817 | 0.912 | 0.861 | **0.846** |
| *CHF* | 0.887 | 0.941 | 0.852 | 0.918 | 0.887 | **0.897** |
| *Pneumonia* | 0.823 | 0.839 | 0.804 | 0.943 | 0.89 | **0.859** |
| *Average AUC* | **0.847** | **0.863** | **0.824** | **0.924** | **0.879** | **0.867** |

With an average AUC of 0.867, this model performs on par with the model that took in only chest X-ray images. However, a vast improvement in Pneumonia classification is observed (0.859 vs. 0.789). In contrast to the image-based models that performed poorly on the Pneumuina class, the text-based model performs significantly better in detecting this class. Almost a 7% AUC increase is present, compared to the model that took in only the chest X-ray as input. The reason the overall average AUC did not increase compared to the chest X-ray model is because of slight performance decreases on average in the Normal and CHF classifications.

4.1.5 Radiology Report: Full Report

The results of 5-fold cross-validation of the model with the whole parts of the reports as input using the architecture shown in Figure 4 can be found in Table 5.

*Table 5: Classification AUC with full report text input*

|  | Fold 1 | Fold 2 | Fold 3 | Fold 4 | Fold 5 | Average |
|---|---|---|---|---|---|---|
| **Normal** | 0.99 | 0.996 | 0.991 | 0.986 | 0.995 | **0.991** |
| **CHF** | 0.967 | 0.991 | 0.963 | 0.941 | 0.952 | **0.963** |
| **Pneumonia** | 0.977 | 0.978 | 0.924 | 0.943 | 0.911 | **0.951** |
| **Average AUC** | **0.978** | **0.988** | **0.959** | **0.957** | **0.952** | **0.967** |

With an average AUC of 0.967, this model is overwhelmingly predictive. It would seem to imply that the diagnosis that radiologists arrive at after doing their reading are very accurate.

4.1.6 Radiology Report Text and Chest X-Rays

The results of 5-fold cross-validation of the model with X-ray and full radiology reports as input using the architecture shown in Figure 6 can be found in Table 6.

*Table 6: Classification AUC with full report text and chest X-ray input*

|  | Fold 1 | Fold 2 | Fold 3 | Fold 4 | Fold 5 | Average |
|---|---|---|---|---|---|---|
| **Normal** | 0.986 | 0.996 | 0.977 | 0.981 | 0.994 | **0.987** |
| **CHF** | 0.971 | 0.982 | 0.961 | 0.917 | 0.952 | **0.956** |
| **Pneumonia** | 0.984 | 0.965 | 0.946 | 0.954 | 0.927 | **0.954** |
| **Average AUC** | **0.981** | **0.981** | **0.959** | **0.951** | **0.958** | **0.966** |

There is no significant difference in classification performance between this model (X-ray and full radiology reports as input) and the model that simply took in the Full Report sentence embedding as input (Table 5). The full report is so powerfully predictive that it dominates the classification performance.

### 4.1.7 Exam Indication Sentence Embeddings and Chest X-Rays

The results of 5-fold cross-validation of the model with X-ray and the indication part of the reports as input using the architecture shown in Figure 6 can be found in Table 7.

*Table 7: Classification AUC with exam indication report text and chest X-ray input*

|  | Fold 1 | Fold 2 | Fold 3 | Fold 4 | Fold 5 | Average |
|---|---|---|---|---|---|---|
| **Normal** | 0.912 | 0.957 | 0.904 | 0.871 | 0.947 | **0.918** |
| **CHF** | 0.969 | 0.974 | 0.932 | 0.968 | 0.966 | **0.962** |
| **Pneumonia** | 0.89 | 0.917 | 0.772 | 0.828 | 0.898 | **0.861** |
| **Average AUC** | **0.924** | **0.949** | **0.869** | **0.888** | **0.937** | **0.913** |

Using the model with X-ray and the indication part of the reports as input, a 5.7% AUC increase is observed compared to the models that took in just the chest X-ray as input (Table 1), and a 4.6% AUC increase compared to the model that took just the Exam Indication sentence embedding as input (Table 4). The model combines the strength of classification of the Normal and CHF class from the chest X-ray model, with the strength of Pneumonia classification from the sentence embedding model. Although this is not as good as the Full Report sentence embedding model (Table 5), it is only dependent on exam indication, which is agnostic to possible radiologist error and biases reflected in the full report.

### 4.1.8 Static Eye-Gaze Heatmap and Chest X-Rays

The results of 5-fold cross-validation of the model with X-ray and static heatmap input using the architecture shown in Figure 7 can be found in Table 8.

*Table 8: Classification AUC with chest X-ray and static heatmap ground truth*

|  | Fold 1 | Fold 2 | Fold 3 | Fold 4 | Fold 5 | Average |
|---|---|---|---|---|---|---|
| **Normal** | 0.891 | 0.944 | 0.914 | 0.885 | 0.91 | **0.909** |
| **CHF** | 0.904 | 0.941 | 0.881 | 0.893 | 0.879 | **0.899** |
| **Pneumonia** | 0.811 | 0.81 | 0.762 | 0.714 | 0.84 | **0.787** |
| **Average AUC** | **0.868** | **0.898** | **0.852** | **0.831** | **0.877** | **0.865** |

The average overall AUC of 0.865 indicates that this model does not have much of a significant improvement compared to the model that simply took in the chest X-ray as input alone with average AUC of 0.856 (Table 1). A similar trend of struggling to classify Pneumonia, i.e., an average Pneumonia AUC of 0.787, reoccurs, and it seems that the chest X-ray input drives the classification accuracy without the eye-gaze information adding much value.

4.1.9 Temporal Eye-Gaze Heatmaps and Chest X-Rays

The results of 5-fold cross-validation of the model with X-ray and temporal heatmap input using the architecture shown in Figure 8 can be found in Table 9.

*Table 9: Classification AUC with chest X-ray and temporal heatmaps input*

|  | Fold 1 | Fold 2 | Fold 3 | Fold 4 | Fold 5 | Average |
|---|---|---|---|---|---|---|
| **Normal** | 0.882 | 0.913 | 0.902 | 0.893 | 0.899 | **0.897** |
| **CHF** | 0.91 | 0.933 | 0.887 | 0.888 | 0.877 | **0.899** |
| **Pneumonia** | 0.802 | 0.821 | 0.744 | 0.721 | 0.823 | **0.782** |
| **Average AUC** | **0.864** | **0.889** | **0.844** | **0.834** | **0.866** | **0.859** |

As observed with the static heatmap ground truth experiment, there is no significant difference compared to the model that only took in chest X-ray as input as the average AUC of 0.859 shows. The results in Table 8 and 9 illustrates that the eye gaze data does not improve the classification performance.

4.2. Explainability Results

In this subsection, we present the explainability results of the experiments explained in Section 3.3. These results are stated based on the mean and median of the attention overlap on the test dataset, which can be found in Table 10. Furthermore, a visual example for the generated attention maps of each experiment is provided in Table 11.

As it can be seen from Table 10, chest X-ray and Exam Indication with static heatmap ground truth provides the highest explainability (mean attention overlap: 0.13) with a median attention overlap comparable to that of chest X-ray and Full Report with static heatmap ground truth (0.089 vs. 0.093).

As can be observed in the visual examples (Table 11), although integrating the reports can help in improving the classification performance, it cannot significantly affect the quality of the attention maps if the eye-gaze heatmaps are not included as the second ground-truth. Without incorporating these heatmaps, most of the attention of the model would be to the outside of lungs. The attention maps generated by the models in which eye-gaze heatmaps are applied as the second ground-truth, however, can better specify the important parts of the images which have been determined by the bounding boxes.

*Table 10: Attention overlap between the generated attention maps of the developed models and ground-truth bounding boxes*

| Input | Attention Overlap: Mean | Attention Overlap: Median |
|---|---|---|
| Chest X-ray (Figure 10) | 0.0565 | 0.0339 |
| Chest X-Ray with Static Heatmap as Ground Truth (Figure 7) | 0.1009 | 0.087 |
| Chest X-Ray and Exam Indication without Static Heatmap Ground Truth (Figure 11) | 0.0719 | 0.0527 |
| Chest X-Ray and Exam Indication with Static Heatmap Ground Truth (Figure 11) | **0.1326** | **0.0886** |
| Chest X-Ray and Full Report without Static Heatmap Ground Truth (Figure 11) | 0.0846 | 0.063 |
| Chest X-Ray and Full Report with Static Heatmap Ground Truth (Figure 11) | **0.1286** | **0.0933** |

*Table 11: Visual examples of the generated attention maps by the trained models*

| Input | Visual Example |
|---|---|
| Chest X-ray (Figure 10) | 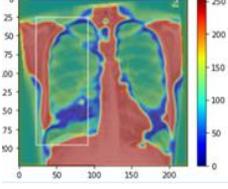 |
| Chest X-Ray with Static Heatmap as Ground Truth (Figure 7) | 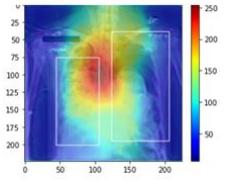 |
| Chest X-Ray and Exam Indication without Static Heatmap Ground Truth (Figure 11) | 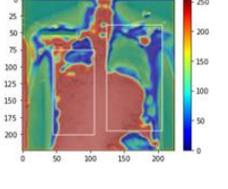 |
| Chest X-Ray and Exam Indication with Static Heatmap Ground Truth (Figure 11) | 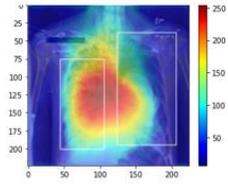 |
| Chest X-Ray and Full Report without Static Heatmap Ground Truth (Figure 11) | 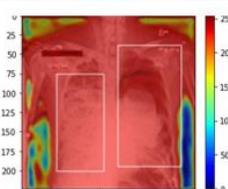 |
| Chest X-Ray and Full Report with Static Heatmap Ground Truth (Figure 11) | 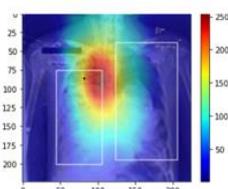 |

## 5. Discussion

5.1 Classification Performance Experiments

Several key observations can be made from the classification performance experiments. Firstly, models taking in visual input such as chest X-ray images struggled to classify the Pneumonia class (e.g., AUC: 0.789). However, sentence embedding based models performed well for that class (e.g., AUC of 0.859 for exam indication text input). Using the Chest X-ray and Exam Indication sentence embedding features as inputs to the classifier resulted in a significant overall classification AUC improvement (from AUC of 85.6% to 91.3%), as Pneumonia was no longer difficult to classify.

Secondly, for the task of Normal vs. CHF vs. Pneumonia classification, the Full Report sentence embedding is highly predictive, with an average AUC of 96.7%. This is the result of radiologists being accurate with their diagnosis in this domain, and this finding may not be extensible to other domains. Nevertheless, as mentioned earlier, even using the Exam Indication section of the report in tandem with the chest X-ray image results in performance improvements. However, the inclusion of eye-gaze information, whether in static or temporal form, did not provide any classification performance improvement.

5.2 Explainability Experiments Discussion

Table 12 shows a summary of the attention overlap for the different input features.

*Table 12: Attention overlap summary for all input feature combinations*

| Input Features | Without Heatmap Loss | With Heatmap Loss | Improvement |
| --- | --- | --- | --- |
| X-Ray | 0.0565 | 0.1009 | 78% |
| X-Ray + Exam Indication | 0.0719 | 0.1326 | 84% |
| X-Ray + Full Report | 0.0846 | 0.1286 | 52% |

There are some interesting trends in the results that are worthwhile to discuss. Visually, the quality of the attention maps improves drastically upon adding the heatmap ground truth to the model architecture. The attention maps learned without the heatmap ground truth are spread over the entire image, and not very focused on key anatomical structures like the heart or lungs. By contrast, the attention maps generated by models using the static eye-gaze heatmap as the secondary ground truth have much cleaner, tighter attention maps that are highly focused on the organs relevant to the diseases being classified.

As is visible from Table 16, these improvements are not merely qualitative. There is a substantial quantitative jump in the attention overlap metric with the inclusion of the eye-gaze heatmaps as the ground truth. The models are able to better highlight relevant portions of the image. What is significant is that these relevant areas marked off by bounding boxes come from a different radiologist than the one whose eye-gaze information is being used to train the models. Thus, the information being extracted by the model from the eye-gaze is valuable and generalizable.

A last observation noticeable from Table 16 is that the models that incorporate text as an input feature alongside the chest X-ray are better able to focus on the relevant area of the image. Thus, text from the radiology report does indeed boost the explainability of the models, though not to the degree of eye-gaze information.

One of the major barriers facing the adoption of DL in medicine is the lack of explainable models. If explainability is made out to be an important goal, then this study concludes that there are significant benefits to collecting eye-gaze information. This information can make models predict based on relevant image areas and improve the explainability of these models. As a result, their entrance into the clinical workflow will be highly facilitated.

The limitation of this research is that the test set for evaluating the explainability of the proposed models only include one class namely Pneumonia. As future work, we will extend our experiments to evaluate the explainability on a test set containing all three classes (Normal, CHF, Pneumonia). In addition, we will conduct a study to collect radiologists' feedback on generated heatmaps, to evaluate whether these heatmaps are radiologically meaningful. This will help to better understand what explainability means to radiologists, which will lead to more useful explainable models.

## 6. Conclusion

In conclusion, it does seem that the best input features for maximizing classification performance are the X-ray image and radiology report itself. The investment to collect eye-gaze information is not worth it as there are no tangible performance gains. However, eye-gaze information is quite beneficial for generating attention maps that better highlight relevant sections of an image, and thus are a worthwhile investment in the pursuit of explainable models. Although adding in the radiology report text did increase the accuracy of the attention maps, the inclusion of the eye-gaze heatmaps as ground truth to tune the attention map generator was significantly more valuable.

2020, pp. 45–55.